# ANALYSIS OF INTEREST POINTS OF CURVELET COEFFICIENTS CONTRIBUTIONS OF MICROSCOPIC IMAGES AND IMPROVEMENT OF EDGES

A. DJIMELI[1], D. TCHIOTSOP[2] and R. TCHINDA[3]

[1] Department of Telecommunication and Network engineering, Laboratoire d'Automatique et d'Informatique Appliquée (LAIA), IUT-FV, University of Dschang
`adjimeli@yahoo.fr`

[2] Department of Electrical engineering, Laboratoire d'Automatique et d'Informatique Appliquée (LAIA), IUT-FV, University of Dschang
`dtchiot@yahoo.fr`

[3] Laboratoire d'Ingénierie des Systèmes Industriels et de l'Environnement (LISIE), IUT-FV, University of Dschang
`ttchinda@yahoo.fr`

*ABSTRACT*

*This paper focuses on improved edge model based on Curvelet coefficients analysis. Curvelet transform is a powerful tool for multiresolution representation of object with anisotropic edge. Curvelet coefficients contributions have been analyzed using Scale Invariant Feature Transform (SIFT), commonly used to study local structure in images. The permutation of Curvelet coefficients from original image and edges image obtained from gradient operator is used to improve original edges. Experimental results show that this method brings out details on edges when the decomposition scale increases.*

*KEYWORDS*

*Curvelet Coefficients, SIFT interest points, Edge Improvement,*

## 1. INTRODUCTION

Edges are discontinuity in the image map resulting from sudden changes of the texture. Contours also mark high frequencies areas in an image. Contours are very important in the characterization of physiognomies and perception. Consequently they are call upon in many applications of image processing such as the analysis of detail in an image, content based retrieval, objects reconstruction and objects recognition.

In practice, contours are determined using filtering operation by applying a convolution of the image with a mask of filter. For this purpose, several masks of convolutions have been defined. Some examples are the filter of Roberts, the filter of Sobel and the filter of Prewitt [1,2]. Unfortunately these filters are sensitive to the noise and they detect contours only in two directions. The Kirsch method uses eight filters to calculate contours in eight different directions, including derivatives in these directions. However the filters shunting devices give sometimes thick contours. By making a convolution of an image with the Laplacian of Gaussian, Marr and Hildreth [2] remove noise which would have been detected by the Laplacian. The success of their method lies in the good choice of the variance σ. Canny et al. proposed an optimal filter for the detection of an ideal contour drowned in a white vibration Gaussien. The filter proposed is optimal in localization and maximizes the signal report ratio on noise [3].

Conventional filtering edges detection techniques operate only in one scale. Multiresolution analysis overcomes many challenges in many domains. Meanwhile separable wavelets are isotropic and could not get edge regularity in an image [4]. This is the main reason why many geometrical transforms have appeared [4,5,6,7]. Curvelet transform [7,8] is non adaptative geometrical transform that has been used with great success in many applications such as denoising, shape recognition, edge detection and enhancement [7,9,10,11,12,13,14,15,16]. Gebäck and Koumoutsakos used the discrete curvelet transform to extract information about directions and magnitudes of features in the image at selected levels of details [14]. The edges are obtained using the non-maximal suppression and hysteresis thresholding of the Canny algorithm. In order to enhance the edges of an image Liu and Qiu have used different means to deal with different scales of the coefficients to enhance the edge of image [15]. The best recent edge detectors through traditional methods start by improving the image quality with Curvelet transform [15,16]. The analysis of Curvelet coefficients contributions in image processing can found in [15,16,18]. Unfortunately authors have not been interested enough on local structure of Curvelet coefficients contributions. To our knowledge, no paper tackles the question of edge improvement of the traditional methods. Meanwhile the potentials of Curvelet decomposition have not been exploited enough. In this paper we analyze local properties of Curvelet coefficients contributions through Scale Invariant Feature Transform (SIFT) [17] and use this study to improve edges in microscopic images.

This paper is organized as follows. In the second section we briefly present Curvelet transform. We describe in section three the coefficients contributions of Curvelet transform in analysing microscopic images. Section four brings out the edge detection algorithm using Curvelet transform and we conclude our work in section five.

## 2. CURVELET TRANSFORM

Two windows are useful to describe Curvelet [7]. Let us consider in continuous domain $\mathbb{R}^2$, $W(r)$ the radial window with value $r \in [1/2, 2]$ and $V(r,t)$ the angular window with value $t \in [-1,1]$. Admissibility conditions are defined by (1) and (2), where $j$ is a radial variable and $l$ is an angular variable.

$$\sum_{j=-\infty}^{+\infty} W^2(2^j r) = 1, \; r \in \left[\frac{3}{4}, \frac{3}{2}\right] \tag{1}$$

$$\sum_{l=-\infty}^{+\infty} V^2(t-l) = 1, \; t \in \left[-\frac{1}{2}, \frac{1}{2}\right] \tag{2}$$

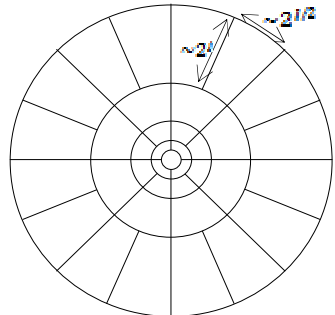

Figure 1. Curvelet tiling of space and frequency where lengths and widths obeying the scaling law $width \approx length^2$

A frequency window $U_j$ is defined by (3) for $j \geq j_0$ in Fourier domain. The Curvelet transform represents a curve as a superposition of functions of various lengths and widths obeying the scaling law $width \approx length^2$. An example of windowing is represented in figure 1 where circular wedge depends on the scale and the direction.

$$U_j(r,\theta) = 2^{3j/4} W(2^{-j}r) V(\frac{2^{\lfloor j/2 \rfloor}\theta}{2\pi}) \tag{3}$$

with $\lfloor j/2 \rfloor$ being the integer part of $j/2$.

Curvelet real values can then be computed as:

$$U_j(r,\theta) + U(r,\theta+\pi) \tag{4}$$

frequency domain Curvelets are supported near a parabolic wedge. However rotation is not adapted to Cartesian array. That is why Candès and Al indexed the values to the origin. Figure 2. (a) shows grey data in the upright parallelogram broken and indexed in a rectangle at the origin.

For Curvelet digitalization, they redefine a new wedge $\tilde{W}_j$ based on concentric squares as shown in figure 2. (b) and expressed by (5) [8].

$$\tilde{W}_j(\omega) = \sqrt{\Phi^2_{j+1}(\omega) - \Phi^2_j(\omega)}, \quad j \geq 0 \tag{5}$$

Where $\Phi$ is defined by the product of two monodimensional bypass windows.

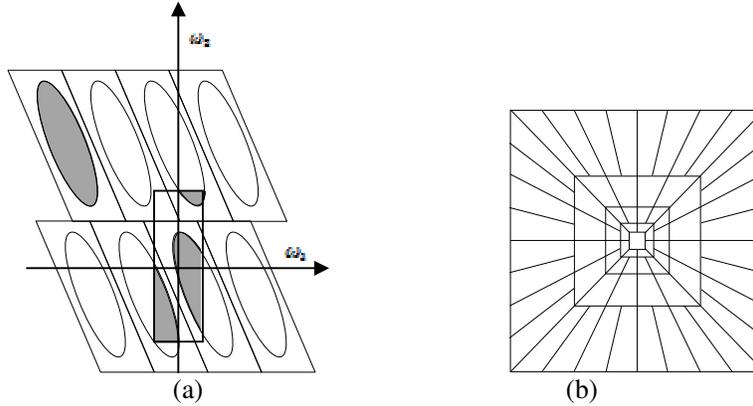

(a)  (b)

Figure 2. (a) Indexation of value to the origin in Cartesian domain, (b) digital tailing of Curvelet. $\Phi$ is useful to separate scales in Cartesian domain. Thus in discrete curvelet Transform, radial and angular windows are obtained as follows:

$$V_j(\omega) = V(\frac{2^{\lfloor j/2 \rfloor}\omega_2}{\omega_1}) \tag{6}$$

$$\tilde{U}_j(\omega) = \tilde{W}_j(\omega) V_j(\omega) \tag{7}$$

The algorithmic structure in [8] is then the application of Fourier transform to the original image to have Fourier coefficients $\hat{f}[n_1, n_2]$, $-n/2 \leq n_1, n_2 < n/2$. For each scale $j$ and a given angle $l$, the product $\tilde{U}_{j,l}[n_1,n_2]\hat{f}[n_1,n_2]$ is then computed and wrap to the origin to obtain (8)

$$\tilde{f}[n_1, n_2] = W(\tilde{U}_{j,l}\hat{f})[n_1, n_2] \qquad (8)$$

The Algorithm ends by computing the inverse Fourier transform of each $\tilde{f}_{j,l}$ to collect Curvelet coefficients $C(j,l,k_1,k_2)$.

In [8] are proposed two methods for implementations of Fast Discrete Curvelet Transform (FDCT): Curvelet via Unequally Spaced Fast Fourier Transform (USFFT) and Curvelet via Wrapping. We chose to implement FDCT via wrapping in MATLAB environment since this method has been proved to be faster.

## 3. ANALYSIS OF CURVELET COEFFICIENTS CONTRIBUTION IN MICROSCOPIC IMAGES

### 3.1 Curvelet coefficients contributions

The aim of microscopic images and many medical images is to extract useful information to underline pathology. For given image of size $m*n$, we apply the Curvelet transform at different scales and appreciate the coefficients contributions. Let $C\{j\}\{l\}(k_1,k_2)$ be a Curvelet coefficient where $j$ is the scale, $l$ the direction parameter and $k = (k_1, k_2) \in \mathbf{Z}^2$. Given the decomposition scale $j$, the coefficients $C\{1\}\{1\}(k_1,k_2)$ are the low frequencies contributions and the other coefficients where $2 \leq j \leq J$ are high frequencies contributions. At a scale $j \geq 2$, we have $N$ orientations such that:

$$N = N_\theta (2^{\lceil (j-2)/2 \rceil}) \qquad (9)$$

Where $N_\theta = 2^P$ is orientation number à scale $j = 2$, $\lceil (j-2)/2 \rceil$ represent the integer part in excess of $(j-2)/2$ and $P \geq 3$.

The Curvelet coefficient contributions in various scales are shown in fig. 3. The original image is $283*275$ and can be decomposed at maximal scale $J = 5$. We can observe that low frequency scale separate darker region from white on and can be useful for watershed segmentation. We discover from coarser scales to finer scales, that Curvelet coefficients contributions become thinner. It is thinner property of thinner scales that will be used to improve edges in microscopic images.

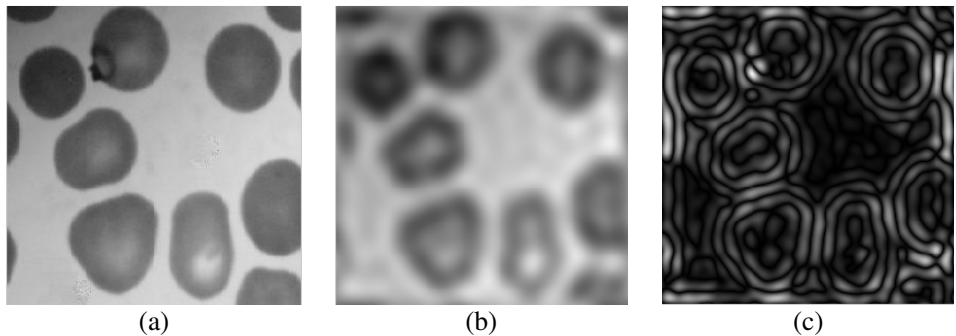

(a)          (b)          (c)

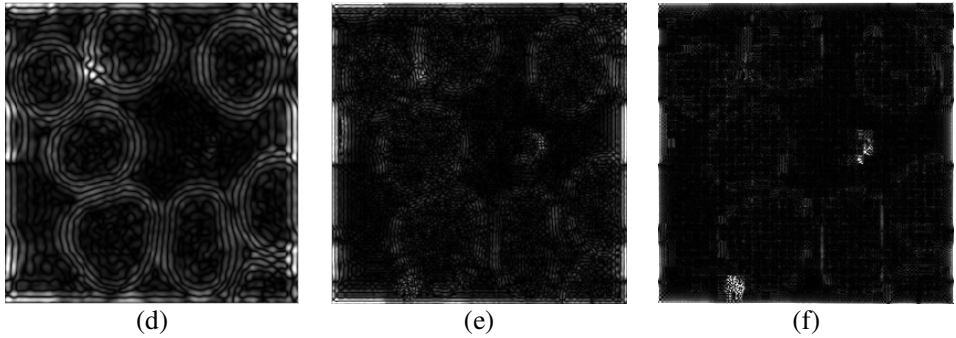

(d) (e) (f)

Figure 3. Curvelet coefficient contributions in each scale: (a) Original image, (b) Curvelet coefficients contributions at thick scale, (c) Enhanced Curvelet coefficients contributions at scale 2, (d) Enhanced Curvelet coefficients contributions at scale 3, (e) Enhanced Curvelet coefficients contributions at scale 4, (f) Enhanced Curvelet coefficients contributions at thinner scale.

## 3.2 Local structure of Curvelet coefficients contributions

The appreciation of a pixel based only on the pixel intensity is not sufficient. Knowing the local structure of a pixel can be useful to appreciate curvelet coefficients contributions. That is why we need to find interest points in order to identify robust characteristics in an image. Many interest points detectors have been presented in literature such as Speeded-Up Robust Features (SURF), Scale Invariant Feature Transform (SIFT), or Harris Detector [17,19]. We focus on SIFT which is the most used. SIFT interest point detector is robust to noise and is invariant to scale change. SIFT algorithm begins by selecting peak in a scale space by calculating minimum and maximum Difference of Gaussian (DoG) through scales, then follows the elimination of the unstable points that are points situated in region with low contrast or points situated on contours. Dominant orientations based on local properties are assigned to interest points to obtain points invariant to rotation. The SIFT algorithm end by assigning to each key point a 128 element vector that can be used for mapping. Algorithmic structure use in this paper is from D. Lowe [20].

Fig. 4 shows with cyan colour interest points magnitudes and directions superimpose on the image. There are 34 interest points in the original image while 61 interest points are found with scale one Curvelet coefficients contributions. The number of interest points through Curvelet scales is gathered in table 1.

Table 1: Number of interest points through Curvelet scales.

| Curvelet coefficients contributions at scale: | 0 (original) | 1 | 2 | 3 | 4 | 5 |
|---|---|---|---|---|---|---|
| Number of SIFT | 34 | 61 | 17 | 5 | 0 | 0 |
| Number of SIFT interest point matching original image | 34 | 17 | 0 | 0 | 0 | 0 |

It clearly appears that the number and magnitude of interest points globally decreases through scales. Some few interest points at the finer scale are due to instability of the local structure of curvelet coefficients contributions and also to the presence of contours.

Signal & Image Processing : An International Journal (SIPIJ) Vol.4, No.2, April 2013

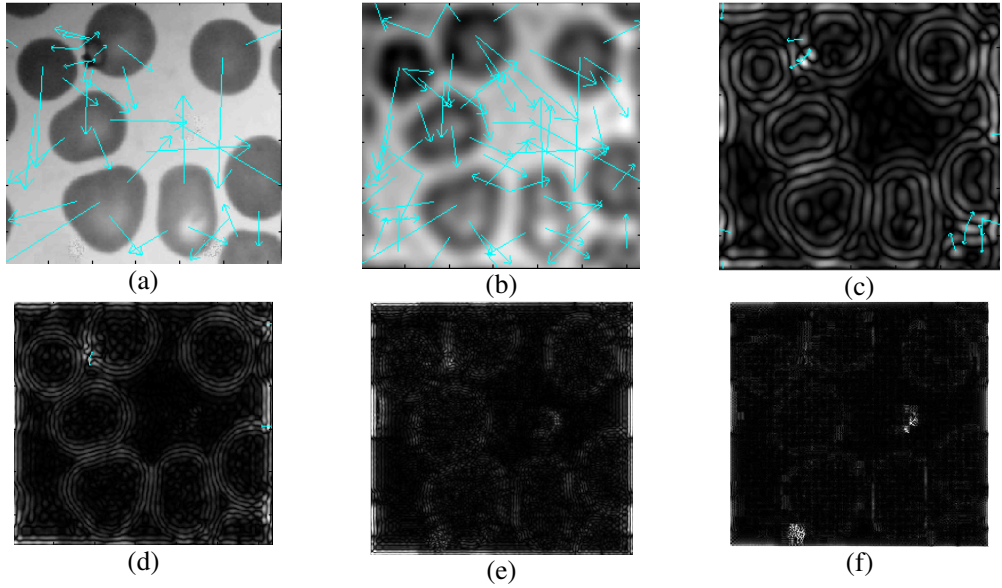

(a)  (b)  (c)

(d)  (e)  (f)

Figure 4. SIFT Interest points of Curvelet coefficients contribution: (a) SIFT Interest points of the original image, (b) SIFT Interest points of the Curvelet coefficients contributions at thick scale, (c) SIFT Interest points of the Curvelet coefficients contributions at scale 2, (d) SIFT Interest points of the Curvelet coefficients contributions at scale 3, (e) SIFT Interest points of the Curvelet coefficients contributions at scale 4, (f) SIFT Interest points of the Curvelet coefficients contributions at thinner scale.

It appears that some few interest points match the original image. These interest points are located on low frequency curvelet coefficients contributions. Those matching interest points are said to be stable through scale and could be useful for quick image data base content retrieval or thick blood smear analysis. Figure 5 present interest points stable through scale.

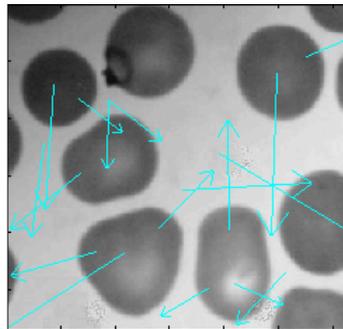

Figure 5. Interest points stable through scales

## 4. EDGE IMPROVEMENT USING CURVELET TRANSFORM

On the contrary of authors of [15,16], who use Curvelet transform to improve the quality of original image before using gradient edge detector to extract contour, we start our algorithm by contour detection and the Curvelet transform is called upon to improve the contour detected.
Canny edge detector [21] was used in our algorithm to obtain first contours. Then follows Curvelet transform of original image and contour image at scale $J$. Thinner scale can be obtained by increasing the size of the image. The contribution of the Curvelet coefficient of the original image at thinner scale then replaces the contribution of the Curvelet coefficient of the contour image. Finally the inverse Curvelet transform of the contour image is computed. Contour is obtained by thinning the image obtained after a threshold operation with two times the mean value. Curvelet transform at scale $J+1$ brings out more precision on contour.

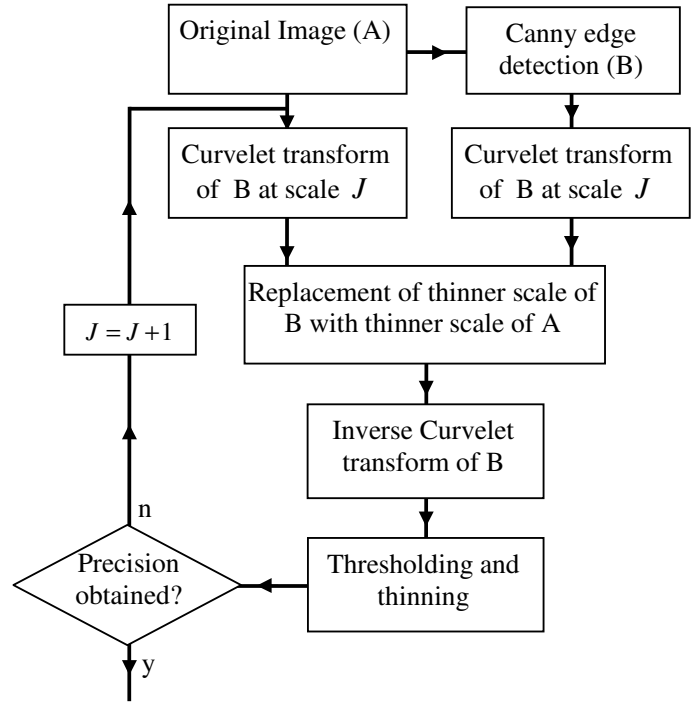

Figure 6. The diagram of our proposed method

An illustration of Edge improvement algorithm using Curvelet is given by the diagram of figure 6. A test example is shown in figure 7 where (a) is our original image filmed in darkness. (b) shows contours obtained with canny edge detector. (c) shows improved edges obtained at scale $J=5$. This image shows contours of glared light in the downright cell that is not present in contours obtained with canny edge detector. Moreover there are new cells contour that appear which are not found on contours from Canny edge detector. (d) depicts improved image obtained at scale $J=6$. This image shows distinct upper left cells contours that are not present in contours obtained with canny edge detector. After series of test with 10 images, we observe that there is no contours improvement at scale $J$ lower than 5. We can conclude that thinner scale $J>4$ can improve edges. The drawback of our method is its execution time because curvelet transform is computed twice.

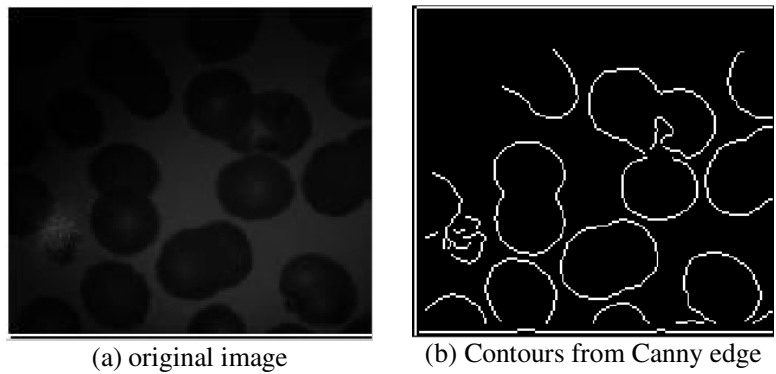

(a) original image      (b) Contours from Canny edge

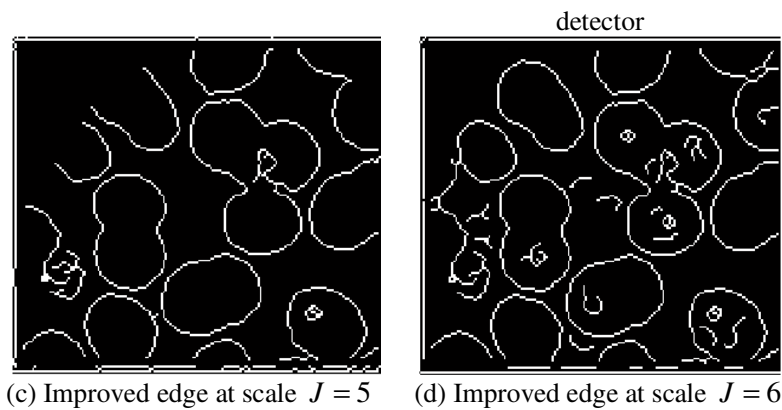

(c) Improved edge at scale $J = 5$    (d) Improved edge at scale $J = 6$

Figure 7. Edge improvement using Curvelet: (a) Original image filmed in darkness, (b) Contours from Canny edge detector, (c) Improved edge at scale $J = 5$, (d) Improved edge at scale $J = 6$

## 5. CONCLUSIONS

This paper presents a method of improvement of contours using the analysis of the coefficients of Curvelet. Each coefficient contribution at specific scale has information that has not yet been capitalized enough. Low frequency scale could help for watershed segmentation and it's SIFT interest point could facilitate content based image retrieval. It is observed that information on contour is located on thinner scale. Curvelet transform is a powerful tool for multiresolution representation of object with anisotropic edge. Experimental results show that thinner scale helps to improve the edges of the images filmed in the darkness.